\documentclass[10pt,twocolumn,letterpaper,pagebackref=true,breaklinks=true,colorlinks,bookmarks=false]{article}

\usepackage{cvpr}
\usepackage{times}
\usepackage{epsfig}
\usepackage{graphicx}
\usepackage{subcaption}
\usepackage{amsmath}
\usepackage{amssymb}
\usepackage{booktabs}
\usepackage[toc,page]{appendix}
\usepackage{multirow}
\usepackage{hyperref}
\usepackage{url}

\newcommand{\nosection}[1]{\vspace{2pt}\noindent\textbf{#1.}}
\cvprfinalcopy % *** Uncomment this line for the final submission

\def\cvprPaperID{****} % *** Enter the CVPR Paper ID here

\begin{document}

%%%%%%%%% TITLE
\title{S2DNAS: Transforming Static CNN Model for Dynamic Inference via Neural Architecture Search}

\author{Zhihang Yuan \thanks{equal contribution}\\
Peking University\\
{\tt\small yuanzhihang@pku.edu.cn}
\and Bingzhe Wu \footnotemark[1]\\
Peking University\\
{\tt\small wubingzhe@pku.edu.cn}
\and Zheng Liang\\
Peking University\\
{\tt\small liangzheng@pku.edu.cn}
\and Shiwan Zhao\\
IBM Research\\
{\tt\small zhaosw@cn.ibm.com}
\and Weichen Bi\\
Peking University\\
{\tt\small biweichen@pku.edu.cn}
\and Guangyu Sun\\
Peking University\\
{\tt\small gsun@pku.edu.cn}
}

\maketitle

\begin{abstract}

  Recently, dynamic inference has emerged as a promising way to reduce the computational cost of deep convolutional neural networks~(CNNs). In contrast to static methods (e.g., weight pruning), dynamic inference adaptively adjusts 
  the inference process according to each input sample, %This property 
  which can considerably reduce the computational cost on ``easy" samples while maintaining the overall model performance.

%shiwan: In this paper, we introduce a general framework, S2DNAS, which enables dynamic inference of various CNN models. 
In this paper, we introduce a general framework, S2DNAS, which can transform various static CNN models to support dynamic inference via neural architecture search. 
To this end, based on a given CNN model, we first generate a CNN architecture space in which each architecture is a multi-stage CNN generated from the given model using some predefined transformations. Then, we propose a reinforcement learning based approach to automatically search for the optimal CNN architecture in the generated space. At last, with the searched multi-stage network, we can perform dynamic inference by adaptively choosing a stage to evaluate for each sample. Unlike previous works that introduce irregular computations or complex controllers in the inference or re-design a CNN model from scratch, our method can generalize to most of the popular CNN architectures and the searched dynamic network can be directly deployed using existing deep learning frameworks in various hardware devices.

\end{abstract}

\section{Introduction}

In the past years, deep convolutional neural networks~(CNNs) have gained great success in many
computer vision tasks, such as image classification~\cite{DBLP:conf/nips/KrizhevskySH12,He2016ResNet,Huang2017DenseNet}, object detection~\cite{Ren2015fasterrcnn,DBLP:conf/cvpr/YOLO16,DBLP:conf/eccv/SSD16}, and image segmentation~\cite{DBLP:conf/iccv/MaskRCNN17,Chen2018DeepLab}. However, the remarkable performance of CNNs always comes with huge computational cost, which impedes their deployment in resource constrained hardware devices. Thus, various methods have been proposed to improve computational efficiency of the CNN inference, including network pruning~\cite{LeCun1990Brain,DBLP:conf/nips/WeightPrune15,Li2017Filter} and weight quantization~\cite{DBLP:conf/icml/GuptaAGN15,Wu2016Quantized,Rastegari2016XnorNet}. Most of the previous methods are static approaches, which use fixed computation graphs for all test samples. 

Recently, dynamic inference has emerged as a promising alternative to speed up the CNN inference by dynamically changing the computation graph according to
each input sample~\cite{Teerapittayanon2016BranchyNet,DBLP:conf/icml/AdaptiveNN17,Figurnov2017Spatially,DBLP:conf/cvpr/LCCL17,Wu2018BlockDrop,Huang2018MSDNet,Liu2018TradeOff,DBLP:conf/iclr/FBS19}. The basic idea is to allocate less computation for ``easy" samples while more computation for ``hard" ones. 
As a result, the dynamic inference can considerably save the computational cost of ``easy" samples without sacrificing the overall model
performance.
Moreover, the dynamic inference can naturally exploit the trade-off between accuracy and computational cost to meet varying requirements (e.g., computational budget) in real-world scenarios. 

To enable the dynamic inference of a CNN model, most previous works aim to develop dedicated strategies to dynamically skip some computation operations during the CNN inference according to different input samples. To achieve this goal, these works attempted to add extra controllers in-between the original model to select which computations are executed. For example, well-designed gate-functions were proposed as the controller to select a subset of channels or pixels for the subsequent computation of the convolution layer~\cite{DBLP:conf/iclr/FBS19,DBLP:conf/cvpr/LCCL17,DBLP:journals/corr/ChannelGatting18}. However, these methods lead to irregular computation at channel level or spatial level, which are not efficiently supported by existing software and hardware devices~\cite{DBLP:conf/nips/StructureSparse16,DBLP:conf/isca/Scalpel17,DBLP:conf/micro/ChannelGating19}.
To address this issue, a more aggressive strategy that dynamically skips whole layers was proposed for efficient inference~\cite{Wu2018BlockDrop,Wang2018SkipNet,DBLP:conf/eccv/ConvnetAIG18}. 
Unfortunately, this strategy can only be applied to the CNN model with residual connection~\cite{He2016ResNet}. 
Moreover, the controllers of some methods comes with a considerable complex structure, which cause the increase of the
overall computational cost in the inference (see experimental results in Section~\ref{sec:experiments}).
% Moreover, These methods need to introduce extra controllers into the computational graph, which is not efficiently supported by the hardware devices~\cite{DBLP:conf/micro/ChannelGating19}.

\begin{figure}[tb]
\centering
    \centering
    \includegraphics [width=1\linewidth]{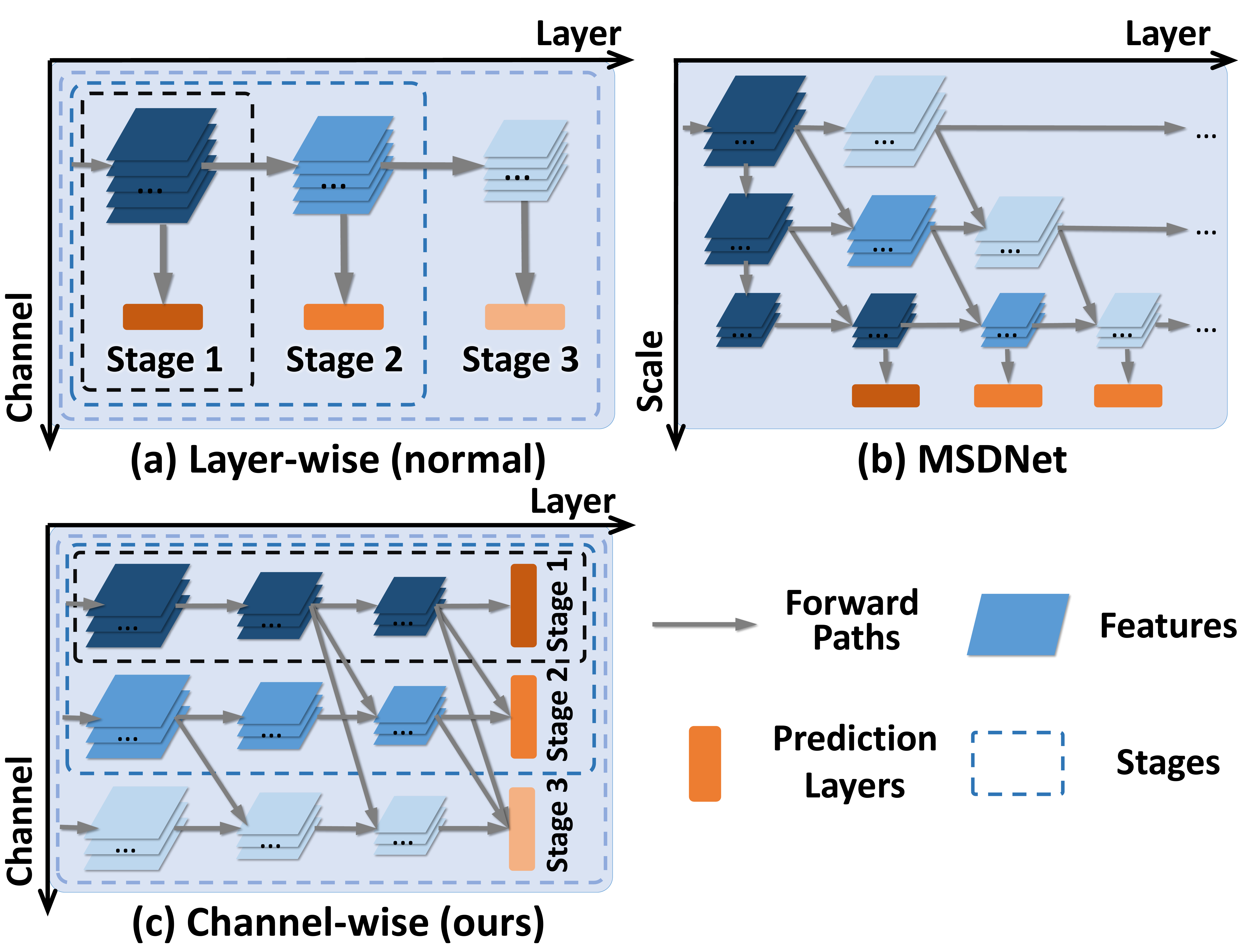}
% \begin{subfigure}[t]{0.5\linewidth} 
%     \centering 
%     \caption{Layer-wise (normal)}
%     \label{fig:naive_early_exit}
%   \end{subfigure}% 
%   \begin{subfigure}[t]{0.5\linewidth} 
%     \centering 
%     \caption{Channel-wise (ours)}
%     \label{fig:ours_paradigm}
%   \end{subfigure} 
\caption{Three paradigms of early exiting methods. (a) The layer-wise approach splits the network into multiple stages along the layer depth. (b) MSDNet devises a multi-stage CNN in which each stage maintains a feature pyramid. (c) Our proposed channel-wise approach splits the network along the channel width.}
\label{fig:introduction}
\end{figure}

To mitigate these problems, researchers propose early exiting the ``easy" input samples at inference time~\cite{DBLP:conf/date/PandaSR16,Teerapittayanon2016BranchyNet,Huang2018MSDNet,DBLP:conf/icann/CascadedInference19a}. A typical solution is to add intermediate prediction layers at multiple layers of a normal CNN model, and then exit the inference when the confidence score of the intermediate classifier is higher than a given threshold. Figure~\ref{fig:introduction}{\color{red}a} shows the paradigm of these early exiting methods~\cite{DBLP:conf/date/PandaSR16,DBLP:conf/icann/CascadedInference19a}. In this paradigm, prediction layers are directly added in-between the original network and the network is split into multiple stages along the layer depth. However, these solutions face the challenge that early classifiers are unable to leverage the semantic-level features produced by the deeper layers. It may cause a significant accuracy drop~\cite{Huang2018MSDNet}. 
%zhaosw: In this paradigm, prediction layers are directly added in-between the original network. As illustrated in Figure~\ref{fig:naive_early_exit}, three prediction layers are added after the $k_1$, $k_2$, $k_3$ convolutional layers. Thus the classifier\footnote{In this paper, the classifier refers to the whole sub-network in the current stage.} in the previous stage cannot use the semantic-level features produced by the classifier in the late stage. 
% In this paradigm, the original network is split into multiple stages along the layer depth, and then the prediction layers are incorporated at different stages accordingly. 
As illustrated in Figure~\ref{fig:introduction}{\color{red}a}, three prediction layers are added to different depth of the network.
% added after the \zhihang{$k_1$, $k_2$, $k_3$ } convolutional layers. 
Thus, the classifier\footnote{In this paper, the classifier refers to the whole sub-network in the current stage.} in the previous stage cannot make use of the semantic-level features produced by the classifier in the late stage. 

Huang et al.~\cite{Huang2018MSDNet} proposed
a novel CNN model, called MSDNet, for solving this issue. The core design of MSDNet is a two-dimensional multi-scale architecture that maintains the coarse and fine level features in every layer as shown in Figure~\ref{fig:introduction}{\color{red}b}.
% the maintenance of the coarse and fine level features in every layer. 
Based on this design, MSDNet can leverage the semantic-level features in every prediction layer and achieve the best result. 
However, MSDNet needs to design specialized network architecture, which cannot generalize to other CNN models and needs massive expertise in architecture design. %As a result, the use of MSDNet can not leverage massive existing knowledge of CNN design for our applications. 

To solve the aforementioned issue without designing CNNs from scratch, we propose to transform a given CNN model into
a \emph{channel-wise} multi-stage network, which comes with the advantage that the classifier in the early stages can leverage the semantic-level features.
Figure~\ref{fig:introduction}{\color{red}c} intuitively demonstrates the idea behind our method. %i.e., the dynamic model generated by our framework follows the paradigm in Figure~\ref{fig:ours_paradigm}. 
Different from the normal paradigm in Figure~\ref{fig:introduction}{\color{red}a}, our method split the original network into multiple stages along the channel width. The prediction layers are added only to the last convolutional layer, thus all classifiers can leverage the semantic-level features. To reduce the computational cost of the classifiers in the early stages, we propose to cut down the number of channels of each layer in different stages (more details can be found in Section~\ref{sec:method}). 

Based on the high-level idea introduced above, we present a general framework called S2DNAS. Given a specific CNN model, the framework can automatically generate the dynamic model following the paradigm showed in Figure~\ref{fig:introduction}{\color{red}c}. 
%To this end, we propose two components to achieve our goal. 
S2DNAS consists of two components: {\tt S2D} and {\tt NAS}. 
First, the component {\tt S2D}, which means ``static to dynamic", is used to generate a CNN model space based on the given model. This space comprises of different multi-stage CNN networks generated from the given model based on the predefined transformations. Then, {\tt NAS} is used to search for the optimal model in the generated space with the help of reinforcement learning. 
Specifically, we devise an RNN to decide the setting of each transformation for generating the model. 
% Note that this controller is only used in the training phase instead of the inference phase, which is different from the previous zhihang{computation skipping} methods.  
To exploit trade-off between accuracy and computational cost, we design a reward function that can reflect both the classification accuracy and the computational cost inspired by the prior works~\cite{Tan2018MnasNet,DBLP:conf/eccv/AMC18,DBLP:conf/cvpr/FBNet19}. 
We then use a policy-gradient based algorithm~\cite{Schulman2017PPO} to train the RNN. The RNN will generate better CNN models with reinforcement learning and we can further use the searched model for dynamic inference. 

%In contrast to previous works, our framework automatically enables the feature interactions between different stages of the generated dynamic model, without the need to manually design the CNN architecture. The inference of the searched dynamic model does not involve extra controllers thus can be directly deployed using the existing deep learning frameworks in various hardware devices. 

To verify the effectiveness of S2DNAS, we perform extensive experiments by applying our method to various CNN models. With a comparable model accuracy, our method can 
achieve further computation reduction in contrast to the previous works for dynamic inference.

%The results demonstrate that our method can significantly reduce the computational cost of the original model without accuracy drop.

\section{Related Work}
\nosection{Static Method for Efficient CNN Inference}
Numerous methods are proposed for improving the efficiency of CNN inference.
Two representative research directions are network pruning~\cite{LeCun1990Brain,DBLP:conf/nips/WeightPrune15,Han2016Compress,Li2017Filter} and quantization~\cite{DBLP:conf/icml/GuptaAGN15,Wu2016Quantized,DBLP:conf/icml/FixedPointQuantize16,Rastegari2016XnorNet}. Specifically, network pruning aims to remove redundant weights in a well-trained CNN without sacrificing the model accuracy. In contrast, network quantization aims to reduce the bit-width of both activations and weights.
Most works in the above two directions are static, which refers to using the same computation graph for all test samples.
Next, we introduce an emerging direction of utilizing dynamic inference for improving the efficiency of CNN inference.

\nosection{Dynamic Inference}
Dynamic inference also refers to adaptive inference in previous works~\cite{DBLP:conf/eccv/ConvnetAIG18,DBLP:journals/corr/ImprovingEarlyExit19}.
% The core idea of dynamic inference is to adaptively reduce the computation operations on ``easy" samples.
Most previous works aim to develop dedicated strategies to dynamically skip some computation during inference. They attempted to add extra controllers to select which computations are executed~\cite{DBLP:conf/cvpr/LCCL17,DBLP:conf/cvpr/SeerNet19,DBLP:conf/cvpr/SBNet18,DBLP:conf/cvpr/NOTALL17,DBLP:journals/corr/ChannelGatting18,DBLP:conf/iclr/FBS19,Wu2018BlockDrop,DBLP:conf/eccv/ConvnetAIG18,Wang2018SkipNet}.
%For example, gate functions were proposed as the controllers for selecting a subset of channels or pixels to skip.
Dong et al.~\cite{DBLP:conf/cvpr/LCCL17} proposed to compute the spatial attention using extra convolutional layers then skipping the computation of inactive pixels. 
Gao et al.~\cite{DBLP:conf/iclr/FBS19} proposed to compute the importance of each channel then skipping the computation of those unimportant channels.
However, these methods lead to irregular computation at channel level or spatial level, which is not efficiently supported by existing deep learning frameworks and hardware devices.
To address this issue, a more aggressive strategy that dynamically skips the whole layers or blocks is proposed~\cite{Wu2018BlockDrop,DBLP:conf/eccv/ConvnetAIG18,Wang2018SkipNet}. 
%The convolutional layers in residual blocks~\cite{He2016ResNet} were skipped according to the output of extra controllers.
For example, BlockDrop~\cite{Wu2018BlockDrop} introduced a policy network  to decide which layers should be skipped.
Unfortunately, this strategy can only be applied to the CNN model with residual connection.
Moreover, these methods introduce extra controllers into the computational graph, the computational cost will remain the same or even increase in some cases. On the other hand, early exiting methods propose to divide a CNN model into multiple stages and exit the inference of ``easy" samples in the early stages~\cite{DBLP:conf/date/PandaSR16,Teerapittayanon2016BranchyNet,Huang2018MSDNet,DBLP:conf/icann/CascadedInference19a}. 
The state-of-the-art is MSDNet~\cite{Huang2018MSDNet} in which the authors manually design a novel multi-stage network architecture to serve the purpose of dynamic inference.

\nosection{Neural Architecture Search}
Recently, neural architecture search~(NAS) has emerged as a promising direction to automatically design the network architecture to meet varying requirements of different tasks~\cite{DBLP:conf/iclr/NAS17,DBLP:conf/cvpr/NASTrans18,DBLP:conf/eccv/AMC18,DBLP:conf/cvpr/AutoDeepLab19,DBLP:conf/cvpr/FBNet19,DBLP:conf/iclr/DARTS19}. There are two typical types of works in this research direction, RL-based searching algorithms~\cite{DBLP:conf/iclr/NAS17} and differentiable searching algorithms~\cite{DBLP:conf/iclr/DARTS19}. In this paper, according to the formulation of our specific problem, we choose the RL-based searching algorithm to search for the optimal model in a design space.

\section{Our Approach}
\label{sec:method}

\begin{figure}[tbp] % h:here 当前位置 % b bottom % t top % p 浮动
    \centering
    \includegraphics [width=1\linewidth]{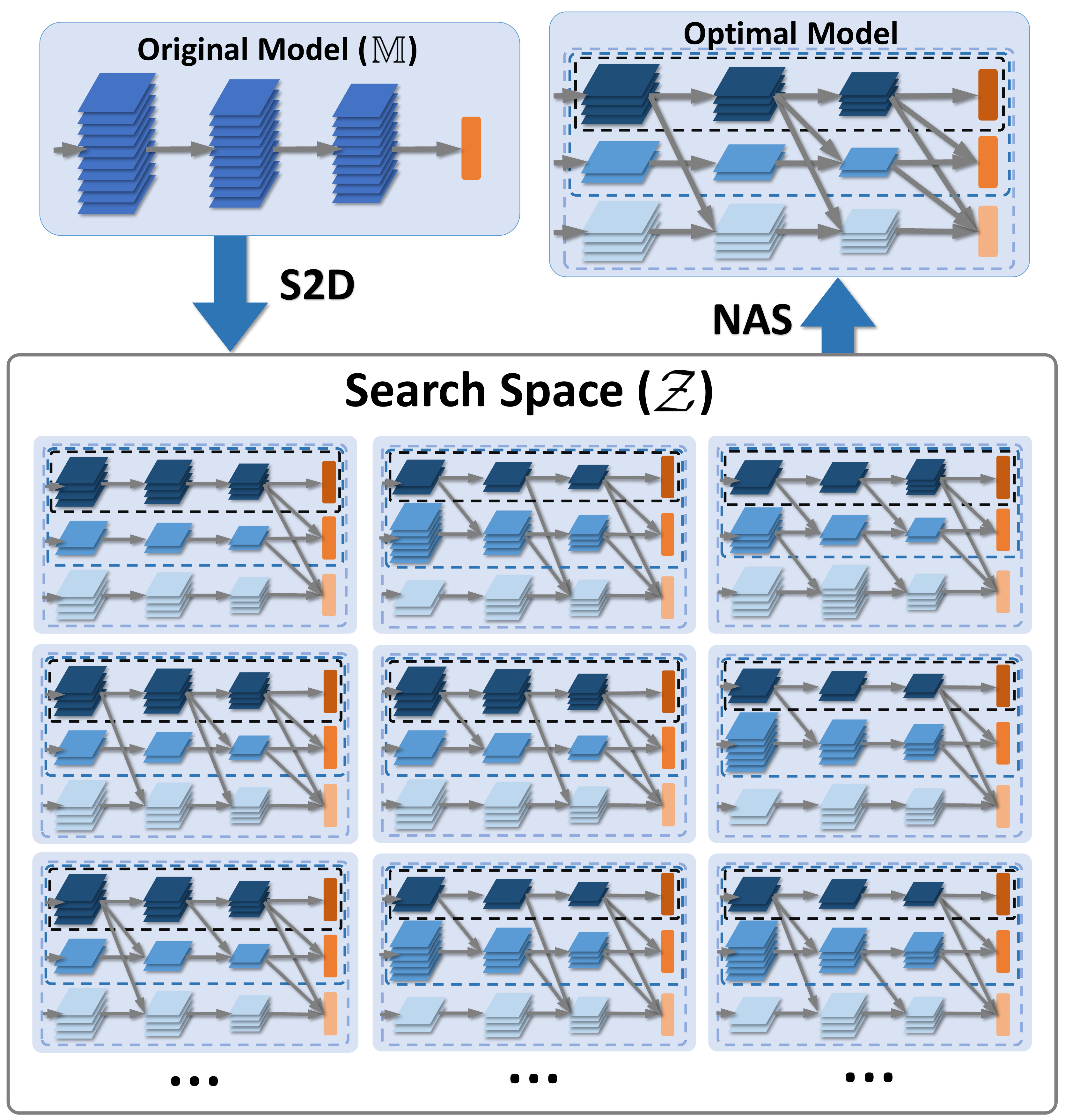}
    \caption{Overview of S2DNAS. {\tt S2D} first generates a search space from the original CNN model. Then, {\tt NAS} searches for the optimal model in the generated space.}
    \label{fig:overview}
\end{figure}

% \begin{figure*}[htbp] % h:here 当前位置 % b bottom % t top % p 浮动
% \centering
%   \begin{minipage}[t]{0.45\linewidth} 
%     \centering 
%     \includegraphics[width=1\textwidth]{fig/overview.pdf}
%   \end{minipage}% 
%   \begin{minipage}[t]{0.52\linewidth} 
%     \centering 
%     \includegraphics[width=1\textwidth]{fig/transformations.pdf}
%   \end{minipage} 
%     \caption{Overview .}
%     \label{fig:overview}
% \end{figure*}

\subsection{Overview of S2DNAS}
The overview of S2DNAS is depicted in Figure~\ref{fig:overview}. At a high level, S2DNAS can be divided into two components, namely, {\tt S2D} and {\tt NAS}. Here, {\tt S2D} means ``static-to-dynamic'', which is used to generate a search space comprises of dynamic models based on a given static CNN model.
Specifically, we define two transformations and then apply the transformations to the original model for generating different dynamic models in the search space. Each of these dynamic models is a multi-stage CNN that can be directly used for dynamic inference. All these generated models form the search space. Once the search space is generated, {\tt NAS} searches for the optimal model in the space. In what follows, we will give the details of these two components.

\subsection{The Details of {\tt S2D}}

\begin{figure}[tb]
    \centering
    \includegraphics [width=1\linewidth]{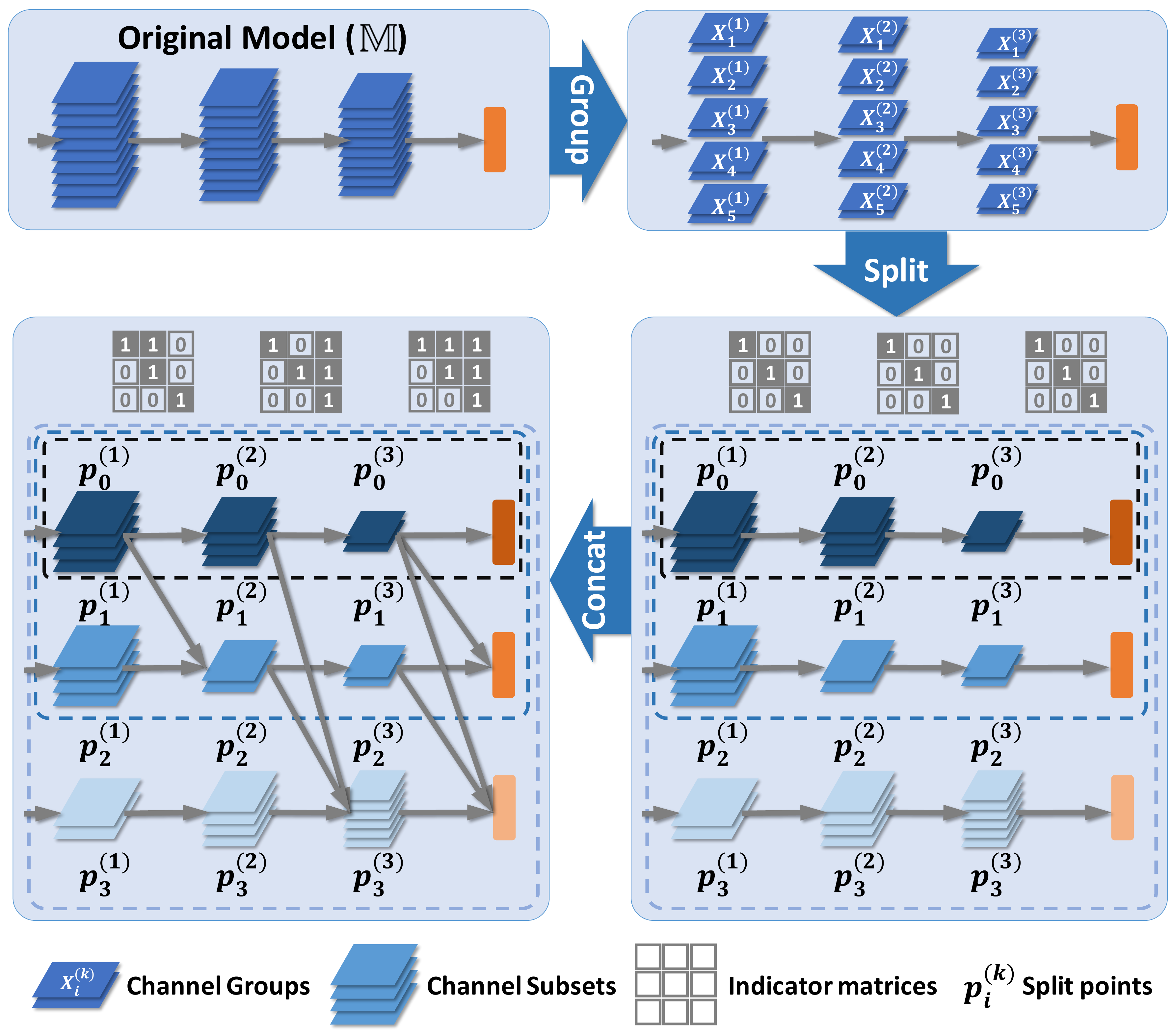}
%    \caption{A concrete example to demonstrate how {\tt split} and {\tt concat} are applied to a three-layers model.}
    \caption{Illustration of how {\tt split} and {\tt concat} are applied to a CNN model. Note that Group is
    an intermediate step of {\tt spilt} for reducing the size of search space (see more details in the main text).}
    \label{fig:trans_example}
\end{figure}

Given a CNN model $\mathbb{M}$, the goal of {\tt S2D} is to generate the search space $\mathcal{Z}$ which consists of different dynamic models transformed from $\mathbb{M}$. Each network in $\mathcal{Z}$ is a multi-stage CNN model in which each stage contains
one classifier. These multi-stage CNNs can be generated from $\mathbb{M}$ using two transformations, namely, {\tt split} and {\tt concat}. First, we propose {\tt split} to split the original model along the channel width as Figure~\ref{fig:trans_example} shows. Specifically, we divide the input channels in each layer of the original model into different subsets. And each classifier can use features from different subsets for prediction. The prediction can be done by adding a prediction layer (shown as yellow squares in Figure~\ref{fig:trans_example}).
Moreover, to enhance the feature interactions between different stages for further performance boost, we propose {\tt concat} to
enforce the classifier in the current stage to reuse the features from previous stages. Next, we will present the details of these two transformations, {\tt split} and {\tt concat}. Before that, we first present some basic notations.

\nosection{Notation} %We start with the computation of a normal convolutional layer.
We start with the notation of a normal convolutional layer. 
Taking the $k$-th layer of a deep CNN as an example, the input of the $k$-th layer is denoted as $\mathbf{X}^{(k)}=\{x_1^{(k)},\cdots,x_C^{(k)}\}$, where $C$ is the number of input channels and $x_i^{(k)}$ is the $i$-th feature map with a resolution of $(h,w)$. We denote the weights as $\mathbf{W}^{(k)}=\{w_1^{(k)},\cdots, w_O^{(k)}\}$,
where $O$ is the number of output channels and $w_i^{(k)}\in \mathbb{R}^{k_c\times k_c\times C}$ ($k_c\times k_c$ is the kernel size). 
In the following parts, we will present two transformations that can be applied to the original model.
The goal of the transformations is to transform a static CNN model $\mathbb{M}$ to a multi-stage model, which can be represented as $a=\{f_1, \cdots, f_s\}$, where $f_i$ is the classifier in the $i$-th stage. 
Next, we will introduce the details of the proposed two transformations.

\nosection{Split} The {\tt split} transformation is responsible for assigning different subsets of the input channels to the classifiers in different stages.
We denote the number of stages as $s$.
A direct way is splitting the input channels into $s$ subsets and allocating the $i$-th subset to the classifier in the $i$-th stage. However, this splitting method results in a considerable large search space which poses the obstacle to the subsequent search process (i.e., {\tt NAS}).
%shiwan: Here, to reduce the search space generated by this transformation, we propose to first divide the input channels into groups and then assign these groups to different stages, i.e., the classifier in each stage will use input channels from several groups.
In order to reduce the search space generated by this transformation, we propose to first divide the input channels into groups and then assign these groups to different classifiers. 

Specifically, we first evenly divide the input channels\footnote{We do not split the input layer.} into $G$ groups thus each group consists of $m=\dfrac{C}{G}$ input channels. Taking the $k$-th layer as an example, this process can be formally denoted as $\mathbf{X}^{(k)}=\{X^{(k)}_1, \cdots, X^{(k)}_G\}$, where $X^{(k)}_i=\{x^{(k)}_{(i-1)m+1},\cdots, x^{(k)}_{im}\}$. 
Once the grouping is finished, these groups are assigned to the classifiers in different stages. 
Precisely, we use the split points $(p_0^{(k)}, p_1^{(k)}, \cdots, p_{s-1}^{(k)}, p_{s}^{(k)})$ to split the groups, here $p_0^{(k)}=0$ and $p_s^{(k)}=G$ are two peculiar points, which denote the start and end points. 
With the split points, we can assign the channel groups $\{X^{(k)}_{p_{i-1}^{(k)}+1}, \cdots, X^{(k)}_{p^{(k)}_{i}}\}$ to the classifier $f_i$ in the $i$-th stage. 
Note that the connection (of the original model $\mathbb{M}$) between different classifiers are removed (see Figure~\ref{fig:trans_example}).

\nosection{Concat} The {\tt concat} transformation is used for enhancing the interaction between different stages. 
The basic idea is to enable the classifiers in later stages to reuse the features from previous stages. Formally, we use indicator matrices $\{\mathbf{I}^{(k)}\}_{k=1}^{L}$ to indicate whether to enable the feature reuse at different positions. 
Here $k$ denotes the $k$-th layer and $L$ is the depth\footnote{Omit the batch normalization and pooling layers.} of the CNN model. 
The element $m_{ij}^{(k)} \in \mathbf{I}^{(k)}$ indicates whether to reuse the features of the $i$-th stage in the $j$-th stage at the $k$-th layer, i.e., $m_{ij}^{(k)}=1$ means that the classifier in the $j$-th stage will concat all the feature maps (of $k$-th layer) from the $i$-th stage. 
Note that we restrict the previous stages from concat the features of the later stages, i.e., $m_{ij}^{(k)}=0, j<i, \forall k<L$. Moreover, we force the $L$-th layer (the prediction layer\footnote{Refer to the last layer of the classifier for prediction.}) to concat the features from all the previous stages.
We demonstrate a concrete example in Figure~\ref{fig:trans_example} to illustrate how to use the above two transformations to reshape a CNN model. 

\nosection{Architecture Search Space} Based on the above two transformations, we can generate the search space by transforming the original CNN model.
Specifically, there are two adjustable settings for the two transformations, splitting points and indicator matrices.
Adjusting the splitting points will change the way to assign the feature groups, which is used for the trade-off between accuracy and computational cost of different classifiers.
For example, we can assign more features to the early stages for improving the model performance on ``easy" samples.
Adjusting the indicator matrices accompanies the change of the feature reuse strategy. To reduce the size of the search space, we restrict the feature layers with the same resolution to use the same split and concat settings in our experiments.
Through changing these two settings, we can generate the search space $\mathcal{Z}$ which consists of different multi-stage models. In the following section, we will demonstrate how to search for the optimal model in the generated space.

\subsection{The Details of {\tt NAS}}
\label{subsec:nas}

Once we obtain the search space $\mathcal{Z}$ from the above procedure of {\tt S2D}, the goal of {\tt NAS} is to find the optimal model $a$ with high accuracy and low computational cost. 
Note that the model is jointly determined by the settings of the above two transformations, i.e., the split points and the indicator matrices. 
With a slight abuse of notation, we also refer the architecture $a$ as these two settings and denote $\mathcal{Z}$ as the space which consists of these different settings.
Thus the optimization goal reduces to search for the optimal settings of the proposed transformations which can maximize our predefined metric (see details in the following section). 

However, searching the optimal setting is nontrivial due to the huge search space $\mathcal{Z}$. For example, in our experiment on MobileNetV2~\cite{Howard2017MobileNet}, the size of the search space is around $10^{11}$.
Motivated by the recent progress in neural architecture search~(NAS)~\cite{DBLP:conf/iclr/NAS17,DBLP:conf/cvpr/NASTrans18,DBLP:conf/eccv/AMC18,Tan2018MnasNet}, we propose to use a policy gradient based reinforcement learning algorithm for searching.
The goal of the algorithm is to optimize the policy $\pi$ which further proceeds the optimal model. 
This process can be formulated into a nested optimization problem:
\begin{equation}
\begin{aligned}
    \arg\max_{\pi} & \mathop{\mathbb{E}}_{a\sim\pi}(R(a,\boldsymbol{\theta}^{*}_a,\mathcal{D}_{val})) \\
    \text{s.t. } & \boldsymbol{\theta}^{*}_a=\arg\min_{\boldsymbol{\theta}_a} \mathcal{L}(a, \boldsymbol{\theta}_a, \mathcal{D}_{train})~\text{,}
\end{aligned}
\label{eq:optimization}
\end{equation}
where $\boldsymbol{\theta}_a$ is the corresponding weights of the model $a$ and $\pi$ is the policy which generates the settings of the transformations. $\mathcal{D}_{val}$ and $\mathcal{D}_{train}$ denote the validation and training datasets, respectively. And $R$ is the reward function for evaluating the quality of the multi-stage model.  

To solve the nested optimization problem in Equation~\ref{eq:optimization}, we need to solve two sub-problems, namely, optimizing $\pi$ when $\boldsymbol{\theta}^{*}_a$ is given and optimizing $\boldsymbol{\theta}_a$ when the architecture $a$ is given.
We first present how to optimize the policy $\pi$ when $\boldsymbol{\theta}^{*}_a$ is given.

\nosection{Optimization of the Transformation Settings}
Similar to previous works~\cite{DBLP:conf/iclr/NAS17,DBLP:conf/cvpr/NASTrans18}, we use a customized recurrent neural network~(RNN) to generate the distribution of different transformation settings for each layer of the CNN model.
Then a policy gradient based algorithm~\cite{Schulman2017PPO} is used for optimizing the parameters of the RNN to maximize the expected reward, which is defined in Equation~\ref{eq:reward}. 
Specifically, the reward in our paper is defined as a weighted product considering both the accuracy and the computational cost:
\begin{equation}
R(a,\boldsymbol{\theta}_a,\mathcal{D}) = \text{ACC}(a,\boldsymbol{\theta}_a,\mathcal{D}) \times \text{COST}(a,\boldsymbol{\theta}_a,\mathcal{D})^\omega ~\text{,}
\label{eq:reward}
\end{equation}
where $\text{ACC}(a,\boldsymbol{\theta}_a,\mathcal{D})$ is the accuracy of the multi-stage model on the dataset $\mathcal{D}$. The $\text{COST}(a,\boldsymbol{\theta}_a,\mathcal{D})$ is the average computational cost over the samples of the dataset $\mathcal{D}$ using dynamic inference.  
For a fair comparison with other works of dynamic inference, we use FLOPs~\footnote{Here, we regard one multiply-accumulate (MAC) as one floating-point operation~(FLOP).} as the proxy of the computational cost. $\omega$ is a hyper-parameter which can be used for controlling the trade-off between model performance and the computational cost.
Next, we will introduce how to solve the inner optimization problem, i.e., optimizing $\boldsymbol{\theta}_a$ on the training dataset when the model $a$ is given.

\nosection{Optimization of the Multi-stage CNN}
The inner optimization problem (i.e., solving for $\boldsymbol{\theta}_a^{*}$) can be solved using the gradient descent algorithm. 
Specifically, we modify the normal classification loss function~(i.e., cross-entropy function) for the case of training multi-stage models.
Formally, the loss function is defined as:
\begin{equation}
\mathcal{L}=\sum_{(x, y)\in D_{train}}\sum_{i=1}^s \alpha_i {\rm CE}(f_i(x,\boldsymbol{\theta}_a),y)~\text{,}
\label{eq:loss}
\end{equation}
Here, {\rm CE} denotes the cross-entropy function. The optimization of the above equation can be regarded as jointly optimizing all the classifiers in different stages. The optimization can be implemented using stochastic gradient descent~(SGD) and its variants. 
We use the optimized $\boldsymbol{\theta}_a$ for assessing the quality of the model $a$ generated by the RNN, which can be further used for optimizing the RNN.
In practice, to reduce the search time, following the previous work~\cite{DBLP:conf/cvpr/NASTrans18}, we approximate $\boldsymbol{\theta}^*_{a}$ by updating it for only several training epochs, without solving the inner optimization problem completely by training the network until convergence.

\begin{figure}[tbp] % h:here 当前位置 % b bottom % t top % p 浮动
    \centering
    \includegraphics [width=0.5\textwidth]{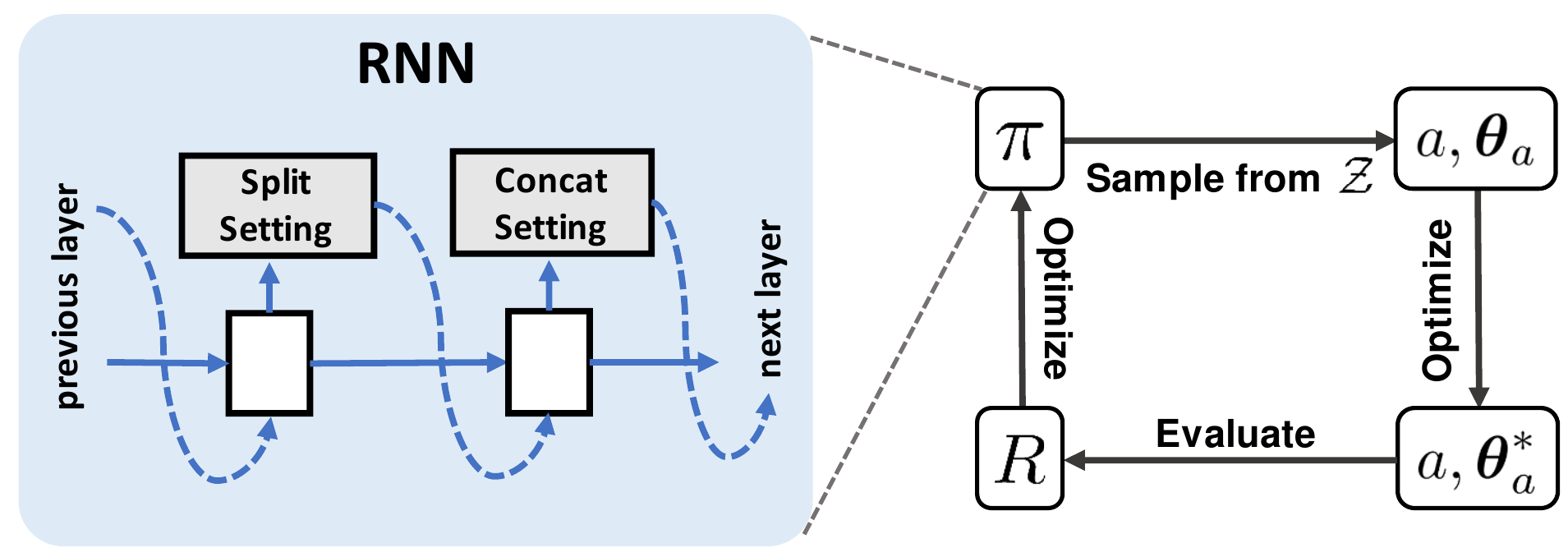}
    \caption{The process of {\tt NAS}. The RNN model is responsible for outputting the policy~$\pi$, i.e., settings of {\tt split} and {\tt concat}, which further produces a model $a$. We can then optimize $a$ for approximating the optimal parameter $\theta_a^{*}$. The computational cost and the accuracy of $a$ is used from evaluating the generated policy $\pi$.}
    \label{fig:nas}
\end{figure}

\nosection{Dynamic Inference of the Searched CNN}

Once the optimal multi-stage model $a=\{f_1, \cdots, f_s\}$ is found, we can directly perform dynamic inference using it. 
Specifically, we set a predefined threshold for each stage. Formally, the threshold of the $i$-th stage is set to $t_i$. 
Then, we can use these thresholds to decide at which stage that the inference should stop. Specifically, given a input sample $x$, the inference stops at the $i$-th stage when the $i$-th classifier outputs a top-1 confidence score $c_i\ge t_i$, here, $c_i = \max(f_i(x))$. %In this paper, we use the constant threshold $t_c$ for all stages, i.e., $t_i=t_c$. 
%Note that the threshold of the classifier should be set before we perform dynamic inference. We propose to use the grid search to find the best threshold. At first, the correctness and confidences of training samples are obtained by inference them on all the $s$ classifiers. Then we set different thresholds of classifiers from the grid of the thresholds and calculate the rewards on training data set using Equation~\ref{eq:reward}. The thresholds with the highest reward are choose.

\section{Experiments}
\label{sec:experiments}

To verify the effectiveness of S2DNAS, we compare it with different dynamic inference methods on different CNN models. 
Our experiments have covered a wide range of previous methods of dynamic inference~\cite{DBLP:conf/cvpr/LCCL17,Wu2018BlockDrop,DBLP:conf/date/PandaSR16,Teerapittayanon2016BranchyNet}. We also evaluate different aspects of S2DNAS, which are presented in the discussion part. 

\begin{table*}[t]
\tiny
\caption{Evaluations on CNN models with different architectures. The number in the Reduction column denotes the relative cost reduction compared with the original model. Some results are missing because there is no implementation of these CNN models in the reference papers.}

\resizebox{\textwidth}{!}{%
\begin{tabular}{cccccccc}
\hline
\multirow{2}{*}{Model}       & \multirow{2}{*}{Method} & \multicolumn{3}{c}{CIFAR-10} & \multicolumn{3}{c}{CIFAR-100} \\
                             &                         & FLOPs & Reduction & Accuracy & FLOPs  & Reduction & Accuracy \\ \hline
\multirow{6}{*}{ResNet-20}   & Original                & 41M   & -         & 91.25\%  & 41M    & -         & 67.78\%  \\
                             & LCCL                    & 30M   & 28\%      & 90.95\%  & 40M    & 1\%       & 68.26\%  \\
                             & BlockDrop               & 45M   & -11\%     & 91.31\%  & 53M    & -29\%     & 67.39\%  \\
                             & Naive                   & 34M   & 18\%      & 91.27\%  & 39M    & 5\%       & 66.77\%  \\
                             & BranchyNet              & 33M   & 20\%      & 91.37\%  & 45M    & -9\%      & 67.00\%  \\
                             & S2DNAS                  & 16M   & 61\%      & 91.41\%  & 25M    & 39\%      & 67.29\%  \\ \hline
\multirow{6}{*}{ResNet-56}   & Original                & 126M  & -         & 93.03\%  & 126M   & -         & 71.32\%  \\
                             & LCCL                    & 102M  & 19\%      & 92.99\%  & 106M   & 16\%      & 70.33\%  \\
                             & BlockDrop               & 74M   & 41\%      & 92.98\%  & 129M   & -2\%      & 72.39\%  \\
                             & Naive                   & 68M   & 46\%      & 92.78\%  & 108M   & 14\%      & 71.58\%  \\
                             & BranchyNet              & 73M   & 42\%      & 92.51\%  & 120M   & 5\%       & 71.22\%  \\
                             & S2DNAS                  & 37M   & 71\%      & 92.42\%  & 62M    & 51\%      & 71.20\%  \\ \hline
\multirow{6}{*}{ResNet-110}  & Original                & 254M  & -         & 93.57\%  & 254M   & -         & 73.55\%  \\
                             & LCCL                    & 166M  & 35\%      & 93.44\%  & 210M   & 17\%      & 72.72\%  \\
                             & BlockDrop               & 76M   & 70\%      & 93.00\%  & 153M   & 40\%      & 73.70\%  \\
                             & Naive                   & 158M  & 38\%      & 93.13\%  & 217M   & 15\%      & 73.06\%  \\
                             & BranchyNet              & 147M  & 42\%      & 93.33\%  & 243M   & 5\%       & 73.25\%  \\
                             & S2DNAS                  & 76M   & 70\%      & 93.39\%  & 113M   & 56\%      & 73.06\%  \\ \hline
\multirow{5}{*}{VGG16-BN}    & Original                & 313M  & -         & 93.72\%  & 313M   & -         & 72.93\%  \\
                             & LCCL                    & 269M  & 14\%      & 92.75\%  & 264M   & 16\%      & 70.46\%  \\
                             & Naive                   & 185M  & 41\%      & 93.34\%  & 202M   & 36\%      & 72.78\%  \\
                             & BranchyNet              & 162M  & 48\%      & 93.39\%  & 239M   & 24\%      & 72.39\%  \\
                             & S2DNAS                  & 66M   & 79\%      & 93.51\%  & 104M   & 67\%      & 72.00\%  \\ \hline
\multirow{5}{*}{MobileNetV2} & Original                & 91M   & -         & 93.89\%  & 91M    & -         & 74.21\%  \\
                             & LCCL                    & 77M   & 15\%      & 93.13\%  & 73M    & 20\%      & 71.11\%  \\
                             & Naive                   & 38M   & 58\%      & 91.90\%  & 61M    & 33\%      & 74.03\%  \\
                             & BranchyNet              & 35M   & 61\%      & 91.76\%  & 74M    & 18\%      & 73.71\%  \\
                             & S2DNAS                  & 25M   & 73\%      & 92.25\%  & 39M    & 57\%      & 73.50\%  \\ \hline
\end{tabular}%
}

\label{table:cifar}
\end{table*}

\subsection{Experiment Settings}

\nosection{Model Setup} In our experiments, we conduct experiments on three CNN architectures: ResNet~\cite{He2016ResNet}, VGG~\cite{DBLP:journals/corr/VGG15}, and MobileNetV2~\cite{MobileNetV218}~\footnote{We use the batch normalization after each convolution layer in VGG and change the stride of the first convolution layer in MobileNetV2 from 2 to 1 for CIFAR.}.
Moreover, to compare with MSDNet, we devised a DenseNet-like model (see more details in the appendix) which has a similar structure
with the MSDNet model. We then perform S2DNAS to the devised model.
We use the same RNN for all our experiments and the details of the RNN is presented in the appendix.

\nosection{Training Details} 
The CIFAR~\cite{CIFAR2009} dataset contains 50k training images and 10k test images. We randomly choose 5k images from the training images as the validation dataset and leave the other 45k images as the training dataset.
We use the same input preprocessing for both CIFAR-10 and CIFAR-100.
To be specific, the training images are zero-padded with 4 pixels and then randomly cropped to 32x32 resolution. The randomly horizontal flip is used for data augmentation.

For the training of the RNN, the PPO algorithm~\cite{Schulman2017PPO} is used. And we use Adam~\cite{DBLP:journals/corr/Adam14} as the optimizer to perform the parameter update in RNN. The details of the hyper-parameters settings can be found in the appendix. 
For the training of the multi-stage model, we use SGD as the optimizer. The momentum is set to $0.9$. The initial learning rate is set to $0.1$ and the learning rate is divided by a factor of $10$ at $50\%$ and $75\%$ of the total epochs. More details of the training settings of different models can be found in the appendix.

For the hyper-parameters of S2DNAS, we set the group number $G=8$ for every layer. And we set the number of stages $s=3$. For comparing with MSDNet which contains 5 stages, we set the $s=5$ for performing S2DNAS on the devised model. The $\omega$ in Equation~\ref{eq:reward} is set to $-0.06$ and all of the $\alpha_i$ in Equation~\ref{eq:loss} is set to $1$ for all experiments. 

\subsection{Classification Results}
In this part, we compare our method with other methods of dynamic inference. To give a comprehensive study of our method, we have covered a wide range of methods, including LCCL~\cite{DBLP:conf/cvpr/LCCL17}, BlockDrop~\cite{Wu2018BlockDrop}, Naive~\cite{DBLP:conf/date/PandaSR16} and BranchyNet~\cite{Teerapittayanon2016BranchyNet}. We conduct experiments on two widely-used image classification benchmarks, CIFAR-10 and CIFAR-100. To show the effectiveness of S2DNAS in reducing the computational cost of CNN models with different architectures, we apply S2DNAS to five typical CNNs with various depth, width, and sub-structures. 

The overall results are shown in Table~\ref{table:cifar}. Note that different thresholds ($t_i$ defined in the previous section) lead to different trade-offs between model accuracy and the computational cost. In our experiments, we chose the threshold which leads to the highest reward on the validation dataset. We also provide further results of using different thresholds in the discussion subsection.  

As shown in Table~\ref{table:cifar}, for most of the architectures and tasks, our method (denoted as S2DNAS in Table~\ref{table:cifar}) can significantly reduce the computational cost with comparable accuracy with the original CNN model.
As mentioned above, we use average FLOPs on the whole test dataset as the metric to measure the computational cost of a given CNN model.
For ResNet-20 on CIFAR-10, S2DNAS has reduced the computation cost of the original net from $41$M to $16$M without the accuracy drop (even with a slight increase as shown in Table~\ref{table:cifar}), which shows a relative cost reduction of $61\%$. 

Our method also shows improvements over other methods for dynamic inference in terms of computational cost reduction. We have reproduced the previous works on these CNN models for comparison. 
We have also implemented a normal early exiting solution (marked as Naive in Table~\ref{table:cifar}), i.e., directly adding prediction layers~(i.e., global average pooling and fully-connected layers) at the intermediate layers of the original models. 
For example, for ResNet-20 on CIFAR-10, compared with BranchyNet~\cite{Teerapittayanon2016BranchyNet}, our method has achieved a slight accuracy improvement (from $91.37\%$ to $91.41\%$) with more computational cost reduction. 

One interesting observation is that some methods even cause an increase in computational cost. For example, BlockDrop boosts the FLOPs of the original net about $29\%$. We infer that this is caused by the controller with high computational cost introduced by BlockDrop in the inference process~\cite{Wu2018BlockDrop}. 
%In contrast, the computation cost of prediction layers in our generated multi-stage models is negligible (lower than 0.1\% of the total computation cost).
We also notice that some of the previous works can not be used for the network without residual connection. 
For instance, BlockDrop cannot be applied to VGG16-BN. In contrast, our method can generalize to CNN without residual connection. From Table~\ref{table:cifar}, our method can reduce the computational cost of the original VGG16-BN net by $79\%$ with a slight accuracy drop. 

\begin{figure}[tbp] % h:here 当前位置 % b bottom % t top % p 浮动
\centering
  \begin{minipage}[t]{0.52\linewidth} 
    \centering 
    \includegraphics[width=1\textwidth]{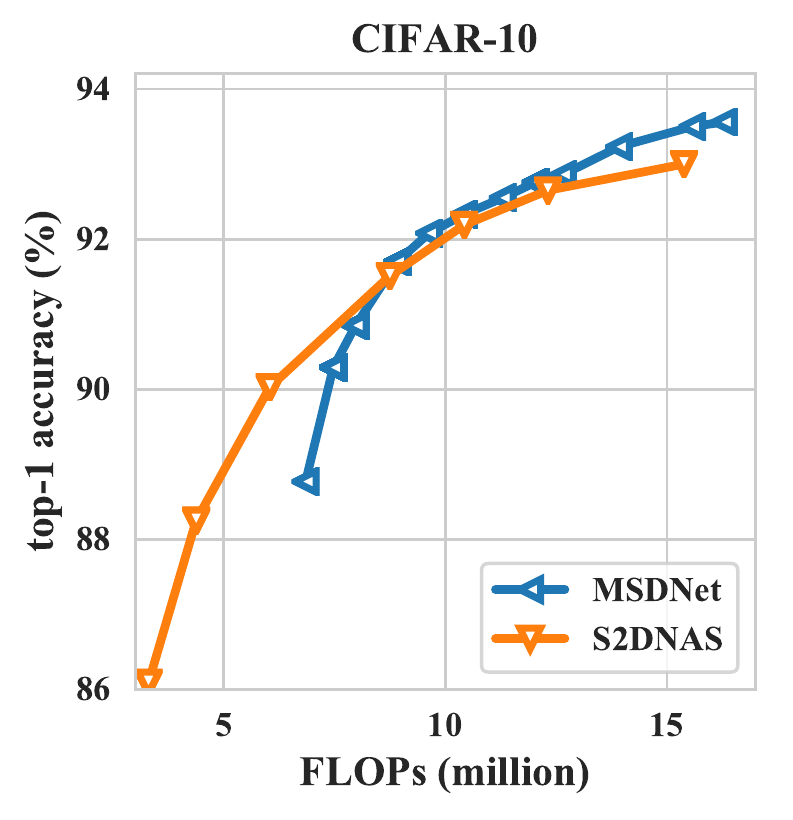}
  \end{minipage}% 
  \begin{minipage}[t]{0.52\linewidth} 
    \centering 
    \includegraphics[width=1\textwidth]{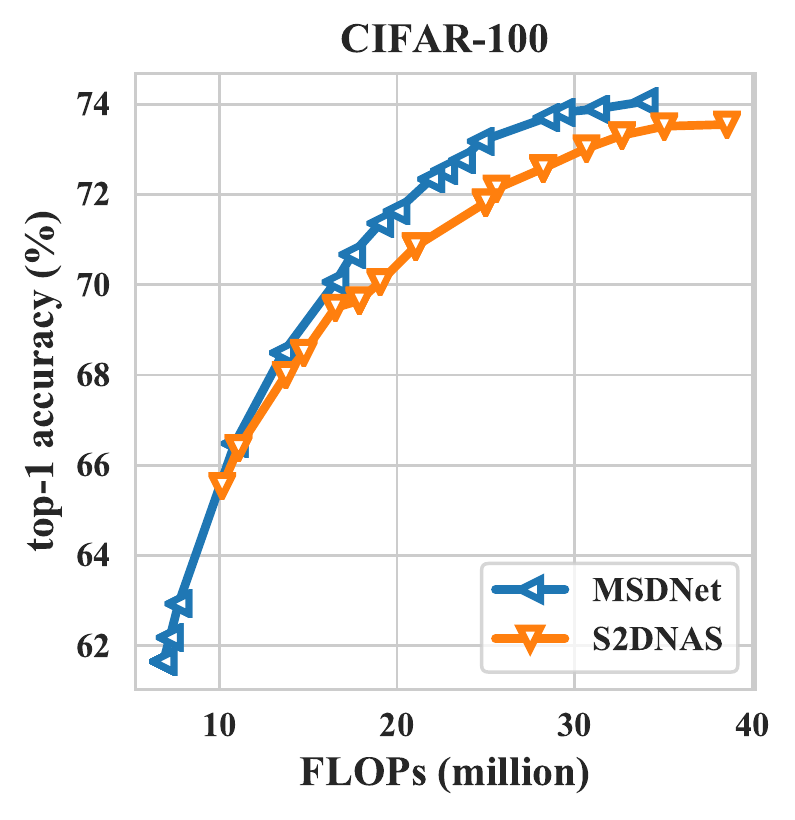}
  \end{minipage} 
    \caption{Comparison to MSDNet.}
    \label{fig:msdnet}
\end{figure}

\nosection{Comparison to MSDNet}
As mentioned in the introduction section, there is a recent work that proposed a specialized CNN named MSDNet for dynamic inference.
Since the method cannot directly be applied to general CNN models, thus for comparison with MSDNet, we design a DenseNet-like~\cite{Huang2017DenseNet} model based on the prior work~\cite{Huang2018MSDNet}, which has similar structure with MSDNet. 
More details of the devised model can be found in the appendix.  
We then apply S2DNAS to it and generate the dynamic models. The results are plotted in Figure~\ref{fig:msdnet}. 
The varying FLOPs metrics of the x-coordinate can be obtained by adjusting the thresholds of each classifier of the dynamic CNN models.
As Figure~\ref{fig:msdnet} shows, in most cases, our method can achieve similar accuracy-computation trade-offs. In the case of CIFAR-10, MSDNet outperforms our method when FLOPs is relative to 15M.
However, the superiority of MSDNet comes with the cost of manually designing the CNN architecture. In contrast, as Table~\ref{table:cifar} shows, our method can be applied to various general CNN models. 

% Then, we will demonstrate the performance of S2DNAS on VGG-16~\cite{VGG14} with batch normalization, DenseNet-40~\cite{DenseNet17} and modified MobileNetV2~\cite{MobileNetV218} for CIFAR. As shown in Table \ref{}, the computation cost of multi-stage networks searched by S2DNAS are decreased by a large margin and this demonstrate our S2DNAS is effective on various network architectures. 
% Note that the DenseNet and MobileNet are both designed for saving computation, so our S2DNAS is orthogonal to the architecture design of the original network. 

% \usepackage{multirow}

\begin{table*}[htbp]
\centering
\tiny
\caption{Accuracy and fractions of samples in test dataset that exit from each stage.}

\resizebox{\textwidth}{!}{%
\begin{tabular}{cccccccc}
\hline
\multirow{2}{*}{Dataset}   & \multirow{2}{*}{Model} & \multicolumn{2}{c}{Stage1} & \multicolumn{2}{c}{Stage2} & \multicolumn{2}{c}{Stage3} \\
                           &                        & Accuracy    & Fractions    & Accuracy    & Fractions    & Accuracy    & Fractions    \\ \hline
\multirow{5}{*}{CIFAR-10}  & ResNet-20              & 98.44\%     & 10.24\%      & 98.89\%     & 41.59\%      & 83.45\%     & 48.17\%      \\
                           & ResNet-56              & 98.25\%     & 67.50\%      & 89.72\%     & 11.19\%      & 75.36\%     & 21.31\%      \\
                           & ResNet-110             & 98.43\%     & 61.66\%      & 93.22\%     & 22.28\%      & 74.28\%     & 16.06\%      \\
                           & VGG-16BN               & 96.54\%     & 87.29\%      & 91.44\%     & 2.22\%       & 68.73\%     & 10.49\%      \\
                           & MobileNetV2            & 98.62\%     & 50.59\%      & 94.04\%     & 33.21\%      & 68.70\%     & 16.20\%      \\ \hline
\multirow{5}{*}{CIFAR-100} & ResNet-20              & 85.27\%     & 58.72\%      & 54.11\%     & 20.68\%      & 29.27\%     & 20.60\%      \\
                           & ResNet-56              & 97.13\%     & 22.27\%      & 86.64\%     & 29.86\%      & 49.51\%     & 47.87\%      \\
                           & ResNet-110             & 95.83\%     & 28.04\%      & 85.05\%     & 28.90\%      & 50.19\%     & 43.06\%      \\
                           & VGG-16BN               & 97.21\%     & 7.18\%       & 90.08\%     & 44.97\%      & 51.20\%     & 47.85\%      \\
                           & MobileNetV2            & 97.81\%     & 8.68\%       & 90.38\%     & 45.84\%      & 51.85\%     & 45.48\%      \\ \hline
\end{tabular}%
}

\label{tab:difficulty}
\end{table*}
\subsection{Discussion}
Here, we present some discussions on our method for providing further insights.

\begin{figure}[tbp] % h:here 当前位置 % b bottom % t top % p 浮动
\centering
  \begin{minipage}[t]{0.52\linewidth} 
    \centering 
    \includegraphics[width=1\textwidth]{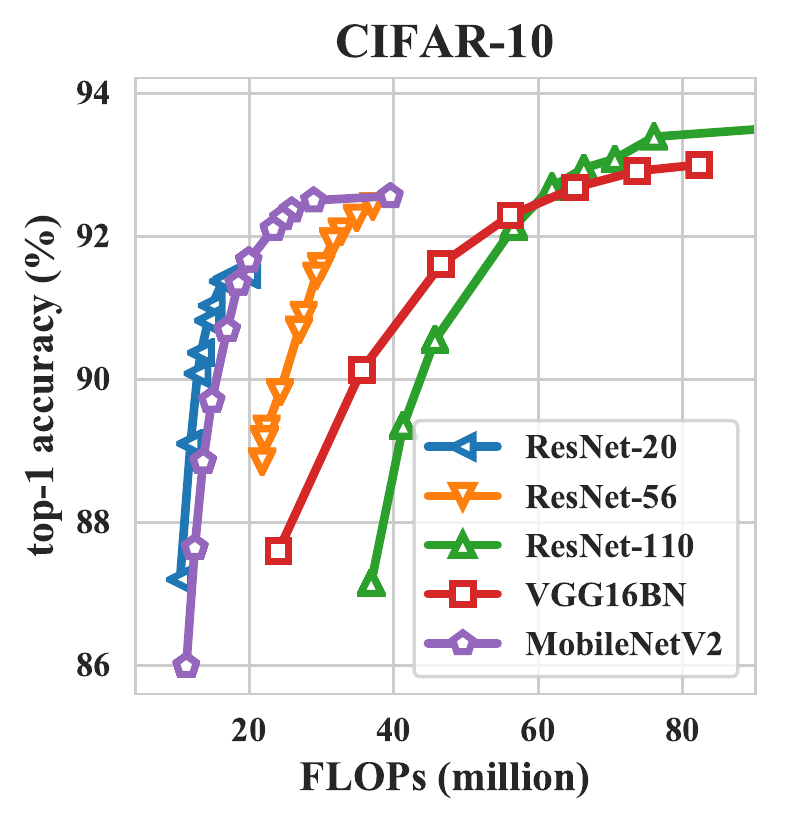}
  \end{minipage}% 
  \begin{minipage}[t]{0.52\linewidth} 
    \centering 
    \includegraphics[width=1\textwidth]{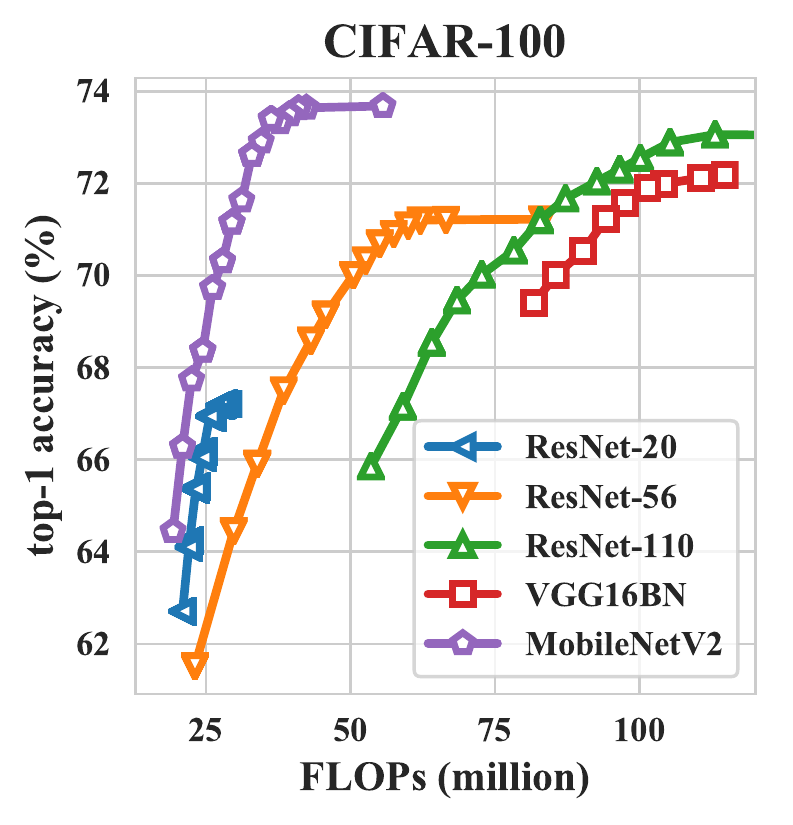}
  \end{minipage} 
    \caption{Trade off accuracy with computational cost by adjusting thresholds of different stages.}
    \label{fig:trade-off}
\end{figure}

\nosection{Trade-off of Accuracy and Computational Cost} A key hyper-parameter of dynamic inference is the threshold setting $t={t_1, \cdots, t_s}$, where $s$ is the number of stages. When the model is trained, different threshold settings lead to different trade-offs between the accuracy and the computational cost. To demonstrate how
the threshold affects the final model performances, we conduct experiments with different thresholds and plot the results in Figure~\ref{fig:trade-off}. All these results show the trend that the increase of computational cost leads to a performance boost. Thus, for practical use, we can set the threshold based on the computational budget of the given hardware device. Moreover, this property also helps to solve the anytime prediction task proposed in the prior work~\cite{Huang2018MSDNet}.

\nosection{Difficulty Distribution of Test Dataset} The basic idea of our method is early exiting ``easy" samples from the early stages. 
In this part, we give the statistics of all the samples in the test dataset ($s$=3, i.e., there are three stages in the trained model). As shown in Table~\ref{tab:difficulty}, for ResNet-20 on CIFAR-10/100, the inference process of about $50\%$ test samples exits from the first two stages. As a result, S2DNAS can considerably reduce the average computation cost. Further, we observe that the accuracy of the classifier in the first stage\footnote{Here, we only consider samples that exit from this stage.} is much higher than the classifier in later stages, which indicates that the classifier can easily classify those samples, i.e., those samples are ``easy" samples. 
This observation also validates the intuition (``easy" samples can be classified using fewer computations) pointed out by some recent works~\cite{DBLP:conf/icml/AdaptiveNN17,DBLP:conf/cvpr/LCCL17,Huang2018MSDNet,DBLP:conf/iclr/FBS19}.

%\nosection{Knowledge Distillation for Improving Training}
%A recent work~\cite{Huang2018MSDNet} has demonstrated that knowledge distillation can improve the training efficiency of a multi-stage CNN. The core idea is to train the network in the early stages by adding supervision signals from the network in the last stage. They took MSDNet as an example to demonstrate the effectiveness of the knowledge distillation technique. In our experiments, we also attempt to incorporate this trick into our framework. However, we found that knowledge distillation can not improve the training in our case. We infer it is caused by the challenge of balancing the training of the original net and the RNN controller. In future work, we will attempt to carefully design the distillation strategy to improve our method.

%We can adjust the thresholds for stages of the multi-stage network to trade the accuracy with computation cost to meet the dynamic computation cost constraint for real-world applications. We change the $\omega$ in our reward function and using the method of grid search in Section 3.6 to get best thresholds. Then, we evaluate the accuracy and the average computation cost on the test dataset and the results is shown as Figure \ref{fig:trade-off}.

\section{Conclusion}
In this paper, we present a general framework called S2DNAS, for transforming various static CNN models into multi-stage models to support dynamic inference. 
%The dynamic model generated using S2DNAS comes with the following advantages
Empirically, our method can be applied to various CNN models to reduce the computational cost, without sacrificing model performance. In contrast to previous methods for dynamic inference, our method comes with two advantages: (1) With our method, we can obtain a dynamic model generated from an existing CNN model instead of manually re-designing a new CNN architecture. (2) The inference of the generated dynamic model does not introduce irregular computations or complex controllers. Thus the generated model can be easily deployed on various hardware devices using existing deep learning frameworks. 

These advantages are appealing for deploying a given CNN model into hardware devices with limited computational resources. To be specific, we can first use S2DNAS to transform the given model into the dynamic one then deploy it on the hardware devices. Moreover, our method is orthogonal to previous pruning/quantization methods, which can further reduce the computational cost of the given CNN model. All these properties of our method imply a wide range of
application scenarios where the efficient CNN inference is desired. 
{\small
\bibliographystyle{ieee_fullname}
\bibliography{main_final}
}

\begin{appendices}

\section{Details of RNN Model and its Optimization}
The RNN model contains a GRU layer with 64 hidden units, $2\times k$ predictors and an embedding layer. The predictors are used to output the probabilities of different transformation settings (split settings and concat settings) for different layers. The number of different split settings in a layer is the combination number $\binom{G-1}{s-1}$, in which $G$ is the number of groups and $s$ is the number of stages. Thus the predictor~(contains a fully-connected layer and a softmax function) is used to predict the probabilities of selecting the $\binom{G-1}{s-1}$ settings. The number of different concat locations is $\frac{s(s-1)}{2}$ and we use the predictor~(contains a fully-connected layer and a logistic function) to predict the probability for each concat locations. Finally, the embedding layer turns the sampled settings of the previous step into dense vectors of fixed size as the input of the GRU layer.

We employ Proximal Policy Optimization~(PPO) to optimize the parameters of the RNN. Adam is used for optimizing the parameters of the RNN model, with a learning rate of 0.001. The number of epochs for PPO is set to 4, the clip parameter is set to 0.1, the mini-batch size is set to 4, the coefficient of value function loss is set to 0.5 and the entropy coefficient is set to 0.01. 

\section{Details of DenseNet-like Model}
To compare with MSDNet, a DenseNet-like model is devised. Specifically, we modify the DenseNet-BC~(k=8, depth=100) by doubling the growth rate after each transition layer and modify the $1\times 1$ convolution in the bottleneck layers of DenseNet by halving the number of output channels. We denote it as DensNet*.

\section{Details of Training Settings}
During network architecture search, we optimize the multi-stage model for 6 epochs using the training dataset to approximate $\boldsymbol{\theta}^{*}_a$. 10k models are sampled from the architecture search space $\mathcal{Z}$ for each experiment. Then the models with top-10 rewards are used to apply the full training. Table~\ref{table:training_settings} demonstrates the hyper-parameters of the full training for different architectures. The scheme of learning rate warm-up is used for 100 iterations.

% \usepackage{graphicx}
\begin{table}[!h]
\centering
\caption{Training Settings}
\label{table:training_settings}
\resizebox{0.5\textwidth}{!}{%

\begin{tabular}{ccccc}
\hline
Model     & Datasets & Batch size & Training epochs & Weight decay \\ \hline
ResNet    & CIFAR    & 128        & 200                                                                              & 1e-4         \\
VGG       & CIFAR    & 64        & 200                                                                              & 5e-4         \\
Mobilenet & CIFAR    & 128        & 300                                                                              & 1e-4         \\
DenseNet* & CIFAR    & 64         & 300                                                                              & 1e-4         \\ \hline
\end{tabular}%
}

\end{table}

\section{Demonstration of a Searched Multi-stage Model}
Table~\ref{table:structure} demonstrates the structure of searched multi-stage ResNet-56 model on CIFAR-10. From this table, we can see that each layer is split into three stages and each stage contains a subset of the original channels. Different stages are concated at different layers thus the feature maps generated by previous layers are reused in the later stages. The accumulated FLOPs increase from 21M to 90M for the three stages. As a result, we can save a lot of computation cost when the "easy" input samples are stopped at stage 1 and stage 2. 

% \usepackage{graphicx}

\begin{table*}[]
\small
\caption{The structure of searched multi-stage ResNet-56 model on CIFAR-10. Building block are shown in brackets, with each element represents the kernel size and the number of output channels.}
\label{table:structure}

\resizebox{\textwidth}{!}{%
\begin{tabular}{cccccl}
\hline
layer name                                                  & output size                    & stage 1                                                                              & stage 2                                                                            & stage 3                                                                              & concat settings                                                                        \\ \hline
conv1                                                       & $32\times 32$                  & $3\times 3$, 6                                                                       & $3\times 3$, 2                                                                     & $3\times 3$, 8                                                                       &                                                                                        \\
\multirow{3}{*}{conv2\_x}                                   & \multirow{3}{*}{$32\times 32$} & \multirow{3}{*}{$\begin{bmatrix}3\times 3, 6 \\3\times 3, 6\end{bmatrix}\times 9$}   & \multirow{3}{*}{$\begin{bmatrix}3\times 3, 2 \\3\times 3, 2\end{bmatrix}\times 9$} & \multirow{3}{*}{$\begin{bmatrix}3\times 3, 8 \\3\times 3, 8\end{bmatrix}\times 9$}   & \multirow{3}{*}{\begin{tabular}[c]{@{}l@{}}stage2-stage1\\ stage3-stage1\end{tabular}} \\
                                                            &                                &                                                                                      &                                                                                    &                                                                                      &                                                                                        \\
                                                            &                                &                                                                                      &                                                                                    &                                                                                      &                                                                                        \\
\multirow{3}{*}{conv3\_x}                                   & \multirow{3}{*}{$16\times 16$} & \multirow{3}{*}{$\begin{bmatrix}3\times 3, 16 \\3\times 3, 16\end{bmatrix}\times 9$} & \multirow{3}{*}{$\begin{bmatrix}3\times 3, 4 \\3\times 3, 4\end{bmatrix}\times 9$} & \multirow{3}{*}{$\begin{bmatrix}3\times 3, 12 \\3\times 3, 12\end{bmatrix}\times 9$} & \multirow{3}{*}{stage2-stage1}                                                         \\
                                                            &                                &                                                                                      &                                                                                    &                                                                                      &                                                                                        \\
                                                            &                                &                                                                                      &                                                                                    &                                                                                      &                                                                                        \\
\multirow{3}{*}{conv4\_x}                                   & \multirow{3}{*}{$8\times 8$}   & \multirow{3}{*}{$\begin{bmatrix}3\times 3, 48 \\3\times 3, 48\end{bmatrix}\times 9$} & \multirow{3}{*}{$\begin{bmatrix}3\times 3, 8 \\3\times 3, 8\end{bmatrix}\times 9$} & \multirow{3}{*}{$\begin{bmatrix}3\times 3, 8 \\3\times 3, 8\end{bmatrix}\times 9$}   & \multirow{3}{*}{stage3-stage2}                                                         \\
                                                            &                                &                                                                                      &                                                                                    &                                                                                      &                                                                                        \\
                                                            &                                &                                                                                      &                                                                                    &                                                                                      &                                                                                        \\
\begin{tabular}[c]{@{}c@{}}prediction\\ layers\end{tabular} & $1\times 1$                    & \begin{tabular}[c]{@{}c@{}}average pool\\ fully-connected\\ softmax\end{tabular}     & \begin{tabular}[c]{@{}c@{}}average pool\\ fully-connected\\ softmax\end{tabular}   & \begin{tabular}[c]{@{}c@{}}average pool\\ fully-connected\\ softmax\end{tabular}     & \begin{tabular}[c]{@{}l@{}}stage2-stage1\\ stage3-stage1\\ stage3-stage2\end{tabular}  \\ \hline
\multicolumn{2}{c}{accumulated FLOPs}                                                        & 21M                                                                                  & 30M                                                                                & 90M                                                                                  &                                                                                        \\ \hline
\end{tabular}%
}
\end{table*}

\end{appendices}

\end{document}